\documentclass[11pt,a4paper]{article}
\usepackage[hyperref]{acl2021}
\usepackage{times}
\usepackage{latexsym}

\usepackage{caption}
\usepackage{subcaption}
\usepackage{graphicx}
\usepackage{booktabs}

\usepackage{amsmath}
\usepackage{adjustbox}

\usepackage{multirow}
\usepackage{longtable}

\usepackage[utf8]{inputenc}
\usepackage{microtype}

\usepackage{array}
\usepackage{xspace}

\usepackage{xcolor,colortbl}
\usepackage{tikz}
\usepackage{collcell}
\usepackage{xstring}
\usepackage{textcomp}

\usepackage{hyperref}

\usepackage{rotating}

\aclfinalcopy 

\newcommand{\vbl}{{\vrule width 1.1pt}}
\newcommand{\hbl}{\noalign{
\hrule height 1.1pt
}}
\newcommand{\pad}{1.4}

\newcolumntype{P}[1]{>{\arraybackslash}m{#1}}

\title{A Discussion on Building Practical NLP Leaderboards:\\The Case of Machine Translation}

\author{Sebastin Santy$^{\dagger}$ \quad
Prasanta Bhattacharya$^{\ddagger}$\\
$^{\dagger}$ Paul G. Allen School of Computer Science \& Engineering, University of Washington\\
$^{\dagger}$  Institute of High Performance Computing (IHPC), A*STAR\\
\texttt{\small ssanty@cs.washington.edu, prasantab@ihpc.a-star.edu.sg} 
}

\date{}

\begin{document}

\maketitle
\begin{abstract}

Recent advances in AI and ML applications have benefited from rapid progress in NLP research. Leaderboards have emerged as a popular mechanism to track and accelerate progress in NLP through competitive model development. While this has increased interest and participation, the over-reliance on single, and accuracy-based metrics have shifted focus from other important metrics that might be equally pertinent to consider in real-world contexts. In this paper, we offer a preliminary discussion of the risks associated with focusing exclusively on accuracy metrics and draw on recent discussions to highlight prescriptive suggestions on how to develop more practical and effective leaderboards that can better reflect the real-world utility of models.
\end{abstract}

\section{Introduction}
Real-world applications of NLP, ML, and AI research in specific contexts like recommendation systems and machine translation, tend to focus primarily on the accuracy and quality of results, which subsequently drive key decisions. However, in the real world, there are various practical constraints that are salient and can pose a trade-off against optimizing accuracy. For example, deploying a model in a high-stakes environment would ideally require the model to be accurate and precise, but also interpretable and explainable. Our recent experiences with machine learning models have taught us that these objectives can often be at odds with each other.

While it is evident that we need to optimize for several metrics, research in building NLP models has largely been driven by benchmarking leaderboards which optimize for only one metric, and more specifically, one accuracy-related metric. Leaderboards have been used extensively for NLU \cite{wang2018glue,wang2019superglue} and QA tasks to \cite{rajpurkar2016squad,nguyen2016ms}. There are popular workshops such as WMT and SEMEVAL which conduct yearly shared tasks and challenges requiring participating teams to design models that can compete on the required metric. These leaderboards are good indicators of model performance based on the chosen metrics, and have drawn considerable attention and participation from the community. However, it is important to point out that the practice of ranking models on just a few metrics does not offer much insight to those who might be interested in deploying these models in real-world applications. In a recent work, \citet{ethayarajh-jurafsky-2020-utility} discuss this utility mismatch between leaderboards (in research) and practice. 

As mentioned earlier, leaderboards are an important tool for tracking and accelerating the progress of the NLP field through the process of gamification, and quantified progress over a given set of metrics. However, and as a result, practitioners who are interested in optimizing for metrics other than accuracy (e.g. energy consumption or model size) have to now manually filter and re-evaluate these models according to their own constraints. It is quite possible, therefore, that practitioners will realize that the models at the top of existing and popular leaderboards tend to underperform on other types of metrics that are more pertinent in the real world. 
This can further widen the gap between academic research and practitioner needs in this field. Due to the way leaderboards are set up, it is often easy to conflate progress in NLP research with adoption and use in real world, which is arguably a more critical consideration. For instance, in certain contexts, machine translation (MT) has claimed to reach human-parity \cite{hassan2018achieving}, although concerns have been raised about the limitations \cite{laubli2018has}. Moreover, and especially for low-resource languages, MT and other multilingual NLP models are far from achieving human parity \cite{joshi2020state}. Despite these limitations, the field has witnessed a competitive show of strength that is driven by the sheer size and availability of both models and data, as pointed out by \citet{bender2021dangers}. For example, even models that were built with a focus on size and efficiency, such as ALBERT \cite{lan2019albert}, have released a much larger counterpart to compete on the leaderboard.

Towards envisioning a better and more sustainable future, we propose that leaderboards can benefit by including as diverse a collection of objectively obtainable metrics as possible. These might include general metrics such as the model size or data summaries, to more task-specific metrics which determine the model quality and usefulness from multiple perspectives, than just accuracy. In this paper, we consider the case of Machine Translation (MT) which is a widely deployed NLP task. We describe the metrics which are commonly used to judge an MT model and understand what factors affect its usability. To support these, we illustrate a few use-cases of how having a portfolio of different metrics, and optimizing for a certain set of them, can be of great value to practitioners in the real world. Based on these observations, we make some recommendations on how future leaderboards can be designed to provide better usability and utility to researchers, practitioners, and end-users.
\begin{table*}[t]
\small
{\renewcommand{\arraystretch}{\pad}
\begin{tabular}{! \vbl >{\columncolor[HTML]{EFEFEF}}l | >{\columncolor[HTML]{EFEFEF}}P{2cm}|P{11cm}|P{1cm}! \vbl }

      \hbl
     \multicolumn{2}{! \vbl l |}{\cellcolor[HTML]{C0C0C0}\textbf{Statistic/Metric}} & \cellcolor[HTML]{EFEFEF}\textbf{Description} & \cellcolor[HTML]{EFEFEF}\textbf{Effort}  \\
      \hbl
      & \textbf{Quality} & \underline{\smash{What is the absolute quality of the translation?}} BLEU \cite{papineni2002bleu} is the most commonly used metric to evaluate the quality of translations on the principles of n-gram overlap. Other similar metrics are METEOR \cite{banerjee2005meteor} and ROUGE \cite{lin2004rouge}. Translation Edit Rate (TER), and especially the Human-targeted Translation Error Rate (HTER), is used to evaluate the number of characters that a human needs to edit to fix the translation in a post-editing setup. & \cellcolor[HTML]{bedbbb} Low\\
      \cline{2-4} 
     &\textbf{Adequacy} & \underline{\smash{Does the output convey the same meaning as the input sentence?}} Along with fluency, this was one of the earliest metrics before automatic evaluation \cite{specia2011predicting}. & \cellcolor[HTML]{ffb6b6} High\\
      \cline{2-4}
     &\textbf{Fluency} & \underline{\smash{How readable is the sentence?}} Even if the content is translated correctly, it can lack the grammatical correctness and idiomatic choices of words that are used. & \cellcolor[HTML]{ffb6b6} High\\
      \cline{2-4}
     &\textbf{Stability} & \underline{\smash{How stable are the outputs during simultaneous interpretation?}} During real-time translation, the translated outputs produced in response to the source sentence should not be inconsistent as they can make the text difficult to parse. It occurs especially when the word-order between source and target language is different \cite{cho2016can}. & \cellcolor[HTML]{ffb6b6} High\\
     \cline{2-4}
     \multirow{-10}{*}{\begin{sideways}MT specific\end{sideways}} &\textbf{Diversity} & \underline{\smash{How diverse are the list of generated translations?}} In a suggestive translation setup, it is important for the MT to be able to generate diverse translations \cite{gimpel2013systematic}. & \cellcolor[HTML]{f4ebc1} Medium\\
      \hbl
      
    &\textbf{Data Size} & \underline{\smash{How much data was the model trained on?}} Previous work has shown how the size of data can affect the performance of different types of MT \cite{koehn2017six}. & \cellcolor[HTML]{bedbbb} Low\\
    \cline{2-4}
    &\textbf{Model Size} & \underline{\smash{What is the size of the model in terms of the number of parameters?}} & \cellcolor[HTML]{bedbbb} Low\\
    \cline{2-4}
    &\textbf{Efficiency} & \underline{\smash{How efficient is the model's training procedure?}} Massive models consume heavy computational resources and can be inefficient to train \cite{strubell2019energy}. & \cellcolor[HTML]{bedbbb} Low\\
    \cline{2-4}
    &\textbf{Latency} & \underline{\smash{What is the time taken to produce a translation output?}} \cite{cherry2019thinking} & \cellcolor[HTML]{bedbbb} Low\\
    \cline{2-4}
    &\textbf{Social Bias} & \underline{\smash{How biased are the outputs of the MT model?}} For instance, gender-bias is a well-known concern with machine translation \cite{stanovsky-etal-2019-evaluating}. & \cellcolor[HTML]{f4ebc1} Medium\\
    \cline{2-4}
    &\textbf{Interpretability} & \underline{\smash{Can we know how the translation was actually carried out?}} \cite{stahlberg-etal-2018-operation} Also, how confident is the MT of the translations it has generated? \cite{bach2011goodness} & \cellcolor[HTML]{f4ebc1} Medium\\
    \cline{2-4}
    \multirow{-9}{*}{\begin{sideways}General\end{sideways}} &\textbf{Robustness} & \underline{\smash{How robust is the MT model}} to adversarial inputs? \cite{cheng-etal-2018-towards} In addition, how well does the system generalize to out-of-domain inputs? \cite{muller2019domain}. & \cellcolor[HTML]{f4ebc1} Medium\\
    
     \hbl

\end{tabular}}
\caption{\small List of statistics and metrics which can be reported for a Machine Translation (MT) model. Some of them are specific to MT, while others are general and applicable to other similar tasks. We describe what each of these metrics mean in the context of MT and also indicate a qualitative estimate of the amount of effort required to measure/compute these statistics/metrics.}
\label{tab:metrics}
\end{table*}

\section{The case of machine translation}
Tasks in NLP can broadly be categorized into two types - applicative and enabling. Applicative tasks are those that have a specific and real-world application, such as text summarization and machine translation (MT). Enabling tasks, on the other hand, are ones that facilitate progress of the field in a certain direction. For example, natural language inference (NLI) can be considered to be an enabling task, for catalyzing progress in the ability to solve logical reasoning problems. For the current discussion, we consider the particular case of MT, which is one of the oldest and most researched fields in AI \cite{hutchins1955georgetown}, as well as one of the most deployed applications of NLP. MT is a sequence-to-sequence task with the objective of translating a sentence from a source to a target language. As a field, MT has evolved from being rule-based, to statistical, to now neural systems.

Since MT is an inherently probabilistic task, where a source sentence can have one of several translations to a target language, researchers have proposed a number of metrics to judge the quality of an MT-generated output. Table \ref{tab:metrics} provides a non-exhaustive list of the most commonly used MT metrics. The quality metrics are a proxy for human judgment scores which are selected based on the high correlation between them. The other metrics capture various aspects of MT which can often be of higher criticality, depending on the kind of application where it is to be applied. It should be noted that not all of these metrics are independent of each other. For instance, inference latency can be strongly linked to the size of the model.

It is evident that an MT model can be judged using several metrics, and optimizing for any one of them might lead to suboptimal performance on the others. While such trade-offs are inevitable in any multi-objective performance optimization scenario, a clear understanding of these trade-offs can better inform model selection in specific contexts. For instance, consider the following four illustrative cases: 

\vskip 0.2cm
\noindent
\includegraphics[height=2\fontcharht\font`\B]{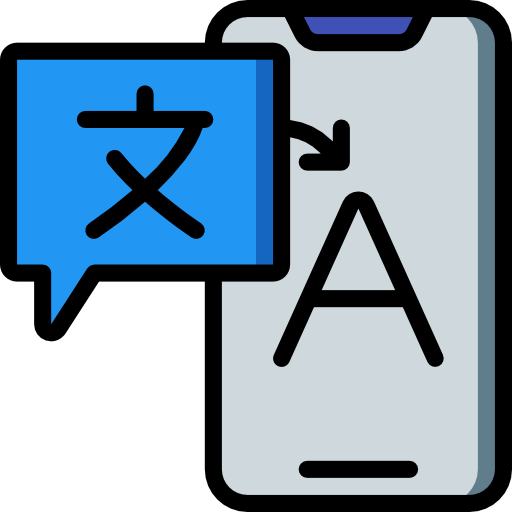}
\underline{Case I.} MT is often used on mobile phones to translate content such as on menu texts, social media and news feed etc. If the MT model is to be deployed locally on the mobile phone, which is a memory-constrained device, then there is a need to optimize the model on {\em model size} and {\em energy efficiency}, in addition to the {\em quality} of translation.
\vskip 0.05cm
\noindent
\includegraphics[height=3\fontcharht\font`\B]{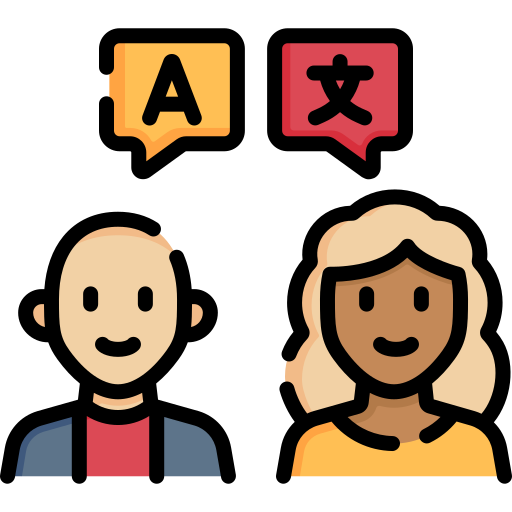}
\underline{Case II.} Real-Time Speech-to-Speech/Text Translation systems are the epitome of current MT research. Many video conferencing systems have started integrating such translation technology e.g., real-time translation from speech to subtitle text. In such cases, it is important to optimize for several metrics such as {\em fluency}, {\em stability}, and most importantly {\em latency}, in addition to {\em quality} of translations.
\vskip 0.10cm
\noindent
\includegraphics[height=2.3\fontcharht\font`\B]{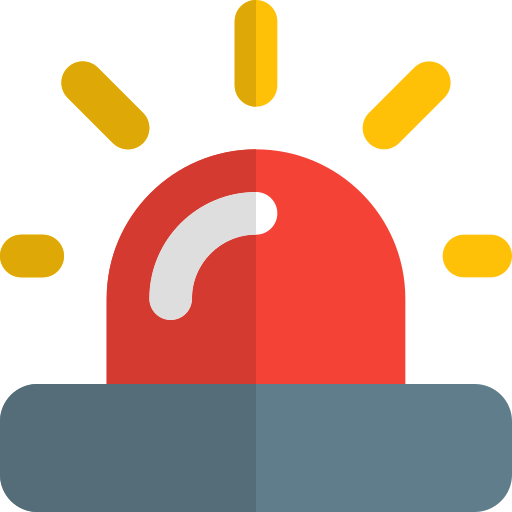}
\underline{Case III.} In critical scenarios, such as while conversing with local law enforcement, or in legal proceedings, the content that is exchanged between parties needs to be {\em adequate}, in terms of the information which it transmits between the source and the target language. Moreover, the translation outputs are required to be {\em interpretable} and should, at the very least, abstain from providing an incorrect or misleading translation, if the translation confidence is low \cite{ong_2017}.
\vskip  0.10cm
\noindent
\includegraphics[height=2.3\fontcharht\font`\B]{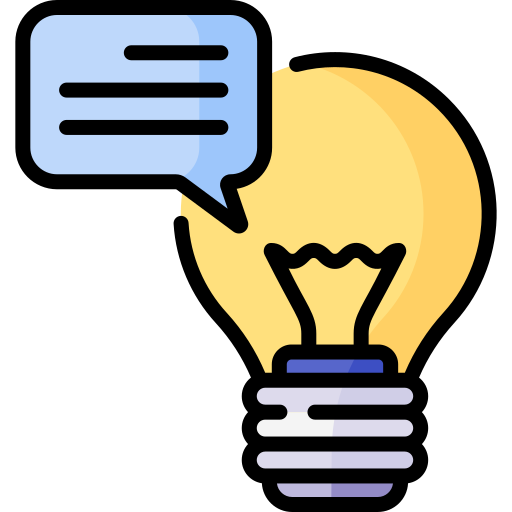}
\underline{Case IV.} MT is often criticized for a lack of pragmatic understanding and contextual reasoning. Hence, MT is often used to assist and complement human translations, through interactive suggestions or cues \cite{knowles2016neural,santy2019inmt}. In such cases, it is important for the generated translations to be low on {\em latency}, and for real-time suggestions to have high {\em diversity} in order to provide multiple options to the human translator to choose from.

\vskip 0.2cm
We acknowledge that the metrics listed in Table \ref{tab:metrics} are not always easy to compute.
For instance, some model-related metrics such as {\em model size}, (training) {\em data size}, {\em model latency} and {\em model efficiency} can be obtained readily, and with relatively low effort. However, measuring {\em social bias}, {\em robustness}, {\em interpretability} and {\em diversity} of the models requires specialized datasets and evaluation procedures, and therefore involves considerable effort. Similarly, metrics such as {\em stability}, {\em fluency} and {\em adequacy} are relatively difficult to compute as they require entirely separate and specialized systems or even human annotators for their evaluations. The amount of effort required to process these statistics and metrics often deters their inclusion in popular leaderboards.











\section{Paths Forward}
In the previous section, we use the case of MT to illustrate the importance of adopting a varied list of model statistics and metrics that can help practitioners with actionable insights and to help them with making more informed choices. In this section, we draw on these observations and other pertinent discussions in the community, to provide some recommendations and considerations on developing effective leaderboards:

\noindent
\textbf{\em i. Multiple metrics in a single leaderboard}\\
Leaderboards for NLP tasks have traditionally reported only single metrics on accuracy/quality. However, and as we have mentioned earlier, some other statistics and metrics such as model size and efficiency are relatively easy to derive, and can be included with little effort in new leaderboards. However, the process of solving NLP tasks often involves solving complex sub-tasks, which now have their own leaderboards and benchmarking datasets such as ones used to measure social bias \cite{stanovsky-etal-2019-evaluating,nadeem2020stereoset,nangia2020crows} and robustness \cite{koh2020wilds,croce2020robustbench}. While these multiple leaderboards help with tracking individual progress, and in mitigating/tackling the prevalent issues in a more structured manner, we believe that it is important to incorporate the varied set of metrics, into a single and primary leaderboard. This would allow the community to better track and evaluate the models on a collection of metrics. Some benchmarks such as EfficientQA \cite{min2021neurips}, Long Arena \cite{tay2020long} have already started incorporating multiple metrics in their efficiency assessment. \citet{gehrmann2021gem} propose a new living benchmark for natural language generation which evolves over time to include new evaluations, metrics and datasets as and when they are released. As a single leaderboard cannot often accommodate all the different metrics, each model can include the model cards \cite{mitchell2019model} as well as data sheets \cite{gebru2018datasheets,bender-friedman-2018-data} which report specific metrics pertaining to the model. As and when the leaderboard evolves to add more metrics, such model cards can help with providing further information about the models.



\noindent
\textbf{\em ii. Fairer leaderboard rankings}\\
Traditionally leaderboards have ranked the models based on a single metric. There has also been evidence of leaderboards averaging accuracy scores over multiple tasks \cite{wang2018glue} or performances on multiple languages \cite{hu2020xtreme}. The process of averaging can, at times, present an incomplete picture of progress. Specifically, the average scores might be driven disproportionately by certain tasks/languages \cite{choudhury2021how}. One simple way to address this is to include several other measures, such as entropy, that can better represent the distribution of the model performance over multiple tasks and languages. Yet another limitation is the leaderboards often tend to become top-heavy and asymptotic around the top, meaning that they do not often reflect the distribution of performances achieved by less capable models.\footnote{Similar and other criticisms are raised by \url{https://dynabench.org/about}} Recent work by \citet{ethayarajh-jurafsky-2020-utility,mishra2021how} suggest that leaderboards should allow users to choose the set of metrics on which to rank the models, rather than providing a rank based on a pre-decided and accuracy-based metric. This strategy has the added benefit of shifting the focus of evaluation from competitively optimizing the accuracy or quality of the system to taking a more holistic approach to solving the task.  Accuracy/Quality metrics do not always reflect model performance in the real-world \cite{laubli2018has,chaganty2018price} and hence competing to optimize a specific metric can often be counterproductive, and can further widen gaps between research and deployability \cite{joshi2019unsung}. This can be mitigated by optimizing on a varied set of metrics, as discussed earlier.




\noindent
\textbf{\em iii. Finding metrics that are useful to users}\\
In this paper, and using the case of MT, we have identified a set of metrics which the research community has discussed in recent studies (Table \ref{tab:metrics}). However, this list is not exhaustive and can include several other metrics along multiple dimensions which can be of concern while deploying a model for a specific purpose. Research often suffers from an echo chamber, and the feedback from the research community often enforces pre-existing norms and practices. We believe that it is important to conduct user studies to collate factors that should be considered during model development and develop appropriate metrics to evaluate models on these factors. It would also be beneficial to constantly update our understanding of how the models would eventually be used. In addition, it is likely that end-users of such deployed applications also have their own set of preferences, for which they optimize. For instance, a person using a maps application might be recommended the shortest route in terms of the time taken or the distance covered. However, the user might want to optimize for a completely different objective, like the aesthetic experience of the journey, or the safety of the route. We contend that leaderboards of the future should afford users the flexibility to define, select and optimize for the kind of metrics that is pertinent to the context at hand.



\noindent
\textbf{\em iv. Community leaderboards and model cards}\\
The current setup of leaderboards allows participants to make a submission, and get their models ranked against other models. However, it can often be difficult to conduct a comprehensive analysis of the models over multiple metrics at the time of submission. The submitted models are often re-run by practitioners in various contexts and using different datasets. Leaderboards can benefit from having a systematic process to update the model cards/leaderboards with these intermediate results from real-world evaluations, so as to increase transparency and reduce repetition of efforts.
\footnote{Similar efforts are being driven by \url{https://huggingface.co/docs}}



\section*{Acknowledgements}
We would like to thank the anonymous reviewers of the Benchmarking workshop as well as the TrustNLP workshop for their valuable feedback and suggestions for improvement. The first author, S.S., benefited greatly from the broad discussions during his graduate school applications, and would like to thank Hal Daumé III, Philip Resnik, Noah A. Smith and Monojit Choudhury in particular for their inputs which helped shape this work. The icons used in this paper were designed by {Pixel Perfect}, {Freepik}, {Smashicons}, from \href{https://www.flaticon.com/}{Flaticon}. 

\bibliography{custom}
\bibliographystyle{acl_natbib}




\end{document}